\documentclass{article}

\usepackage{arxiv}

\usepackage[utf8]{inputenc} % allow utf-8 input
\usepackage[T1]{fontenc}    % use 8-bit T1 fonts
\usepackage{hyperref}       % hyperlinks
\usepackage{url}            % simple URL typesetting
\usepackage{booktabs}       % professional-quality tables
\usepackage{amsfonts}       % blackboard math symbols
\usepackage{nicefrac}       % compact symbols for 1/2, etc.
\usepackage{microtype}      % microtypography
\usepackage{lipsum}
\usepackage{graphicx}
\graphicspath{ {./images/} }
\usepackage{graphicx}
\usepackage[symbol]{footmisc}
\usepackage{subfig}
\usepackage[ruled,vlined]{algorithm2e}
\usepackage{algorithmic}
\usepackage{stackengine}
\usepackage{mathtools}

\usepackage{amsfonts}
\usepackage[export]{adjustbox}
\usepackage[numbers]{natbib}
\usepackage{tabularx}
\usepackage{multirow}
\usepackage{tabularx}
\usepackage{soul}
\usepackage{xcolor}

\title{Rethinking the Function of Neurons in KANs}

\author {
    Mohammed Ghaith Altarabichi\\
    Center for Applied Intelligent Systems Research\\
    Halmstad University, Sweden\\
    mohammed\_ghaith.altarabichi@hh.se
}

\begin{document}
\maketitle
\begin{abstract}
The neurons of Kolmogorov-Arnold Networks (KANs) perform a simple summation motivated by the Kolmogorov-Arnold representation theorem, which asserts that sum is the only fundamental multivariate function. In this work, we investigate the potential for identifying an alternative multivariate function for KAN neurons that may offer increased practical utility. Our empirical research involves testing various multivariate functions in KAN neurons across a range of benchmark Machine Learning tasks.

Our findings indicate that substituting the sum with the average function in KAN neurons results in significant performance enhancements compared to traditional KANs. Our study demonstrates that this minor modification contributes to the stability of training by confining the input to the spline within the effective range of the activation function. Our implementation and experiments are available at: \url{https://github.com/Ghaith81/dropkan}
%Kolmogorov-Arnold Networks (KANs) are recently proposed as an alternative to Multi-Layer Perceptrons (MLPs). KANs rely on learnable activations represented as splines on edges, while KANs' node simply sum the incoming signals. We show in this work that Dropout a popular method to prevent overfitting in MLPs behaves differently during forward and backward passes when applied to KANs. We propose accordingly DropKAN a variant of Dropout designed to regularize KANs by randomly dropping the output of some activations during feedforwad stage instead of dropping nodes.  
\end{abstract}

% keywords can be removed

\section{Introduction}
%Kolmogorov-Arnold Networks (KANs) \cite{liu2024kan} have recently been introduced as an alternative to traditional Multi-Layer Perceptrons (MLPs). KANs differ from MLPs in two key aspects within their computational graph: 1) \textit{Edges}: Instead of the linear weights used in MLPs, KANs employ trainable activation functions. 2) \textit{Neurons (or nodes)}: KANs aggregate incoming signals by summation, in contrast to MLPs which typically apply non-linear activation functions to the inputs. 
This study explores the efficacy of the summation operation in Kolmogorov-Arnold Networks (KANs) nodes. Although opting for the sum operation may appear straightforward given its association with the Kolmogorov-Arnold representation theorem, it is essential to note that the theorem is predicated on only two layers of non-linearity and a small number nodes. Conversely, KANs introduced in \cite{liu2024kan} extend this concept to networks of arbitrary width and depth. This prompts the question of whether the sum operation represents the most optimal choice for KAN nodes in practice. 

In this research, we conducted an empirical study to identify the optimal multivariate function for KANs neurons. We evaluated various candidate functions across a range of machine learning classification tasks to determine which functions yield the best performance. Our findings indicate that using addition as the node-level function may not be ideal, especially for high-dimensional datasets with numerous features. This is because addition can lead to inputs that exceed the effective range of the subsequent activation function, resulting in training instability and reduced generalization performance. To address this issue, we propose utilizing the mean instead of the sum as the node function. Our results show that using the mean helps to maintain inputs within the desired range for activation functions and is consistent with the Kolmogorov-Arnold representation theorem. Furthermore, we investigated the challenge of maintaining inputs within the desired range when utilizing trainable activations in KAN and found that traditional techniques like Layer Normalization \cite{ba2016layer}, often used to tackle covariate shift, may not effectively solve this problem.

%Firstly, we demonstrate that addition is not the most effective function at the node in practical applications, as in practice it could easily lead to an input that falls outside the effective range of the following activation function. Second, we show that the problem of maintaining an input to the spline that falls within the desired range can not be solved with Layer Normalization \cite{ba2016layer}, a commonly used technique to address covariate shift. 
%The remainder of the article is organized as follows. In Section 2, we discuss related work on studying and \hl{tuning} randomness techniques in DNNs. In Section 3, we explain our \hl{hyperparameter optimziation approach} and \hl{define the two proposed new randomization techniques:} loss function noise and gradient dropout. The experiments are presented in Section 4, and the results are further discussed in Section 5. Conclusions are listed in Section 6, while limitations and future work are covered in Section 7. 
\section{Background}
We offer a brief background on Kolmogorov-Arnold representation theorem, and clarify how Kolmogorov-Arnold Networks extends the concept to deeper networks.  
\subsection{Kolmogorov-Arnold representation theorem}
The Kolmogorov-Arnold theorem states that any continuous multivariate function \(f(x_1, \ldots, x_n)\) can be represented as a finite composition of univariate functions, along with the binary operation of addition:
\begin{equation}
f(x_1, \ldots, x_n) = \sum_{q=1}^{2n+1} \Phi_q \left( \sum_{p=1}^n \phi_{q,p}(x_p) \right)
\label{kat}
\end{equation}
where \(\phi_{q,p}\colon [0,1] \to \mathbb{R}\) are continuous univariate functions, \(\Phi_q\colon \mathbb{R} \to \mathbb{R}\) are continuous functions that depend on the sum of \(\phi_{q,p}(x_p)\). This theorem demonstrates that the only fundamental multivariate function is the sum, as all other functions can be constructed using univariate functions and addition. 

\subsection{Kolmogorov-Arnold Networks}
Kolmogorov-Arnold Networks \cite{liu2024kan} defines a KAN layer of   $n_{in}$ inputs and $n_{out}$ outputs as a 1D matrix of functions:
\begin{equation}
\Phi = \{\phi_{q,p}\}, \quad 
      p = 1, 2, \ldots, n_{in}, \quad
      q = 1, 2, \ldots, n_{out}
\end{equation}
%This definition allows to interpret the computation graph described in Equation~\ref{kat} as a two-layers neural network with activation functions \(\phi_{q,p}\) and \(\Phi_{q}\) parameterized with splines and applied to the edges of the inputs and the hidden layer respectively. However, finding the the appropriate functions \(\phi_{q,p}\) and \(\Phi_{q}\) with smooth splines using the shallow network described in Equation~\ref{kat} is impractical, which motivate extending this idea to an arbitrary width and depth by simply stacking more KAN layers similar to deep architectures in neural networks.
This definition enables interpreting the computation graph defined in Equation~\(\ref{kat}\) as a two-layer neural network incorporating activation functions \(\phi_{q,p}\) and \(\Phi_{q}\), which are parameterized with smooth splines and applied to the edges of the inputs and the hidden layer, respectively. However, finding the the appropriate functions \(\phi_{q,p}\) and \(\Phi_{q}\) with smooth splines using the shallow network outlined in Equation~\(\ref{kat}\) proves to be impractical. This challenge prompts the extension of this concept to networks of arbitrary width and depth by simply stacking additional KAN layers, akin to the deep architectures found in traditional neural networks.

%The activation functions \(\Phi\) in a KAN layer are parameterized using splines. 

%The impracticality of finding the appropriate functions with smooth splines is the core motive of generalizing the concept to an arbitary width and depth. 

%to an arbitary width and depth by defining a KAN layer a

\section{Empirical Study and Discussion}
In this empirical study, we investigate nine multivariate functions - sum, min, max, multiply, mean, std, var, median, and norm - applied to ten datasets using a two-layer KAN architecture. The chosen architecture allows for exploring 81 unique combinations of functions between the layers, with each dataset tested across all 81 function settings and ranked based on performance.

Next we detail our experimental design along with the results obtained in the study. In the first experiment we analyze the results of different candidate multivariate neuron functions. In the second experiment, we compare the performance of a KAN network using average (the best function from the previous experiment) in neurons against a regular KAN and KAN equipped with layer normalization across a number of classification problems.

\subsection{Experimental Setup}
\label{Experimental Setup}
Our experiments involve 10 popular~\cite{altarabichi2021surrogate} datasets sourced from the UCI Machine Learning Database Repository\footnote[2]{http://archive.ics.uci.edu/ml}. These datasets encompass a range of sizes and feature diverse domains. In Table~\ref{tab1}, a comprehensive overview of the instances, features, and classes present in each dataset utilized in our research is provided.

The data sets are segmented into training (60\%), validation (20\%), and testing (20\%) partitions. A standardized preprocessing methodology is implemented across all data sets, encompassing encoding of categorical features, handling missing values through imputation, and randomization of instances. Model accuracy serves as the evaluation metric consistently across all experiments. The accuracy scores presented correspond to the performance on the testing set results.

We have chosen to use the default values for the hyperparameters of the KANs model. Furthermore, we have configured the grid to 3 in order to manage the number of parameters effectively. In all experiments conducted on various datasets, we have consistently utilized the KAN architecture [n$_{in}$, 10, 1], where n$_{in}$ represents the number of input features in the dataset. The networks have been trained for 2000 iterations using the Adam optimizer with a learning rate set to 0.01, and a batch size of 32.

%In describing the experiments, we will use KANs with five different settings: using Dropout, DropKAN and neither (No-Drop). For Dropout and DropKAN we will report results in two modes with scale-up (Dropout w/ scale) and without (Dropout w/o scale).

\begin{table*}[t]
\centering
\caption{UCI Data sets used for evaluation.}\label{tab1}
\begin{tabular}{|l|l|l|l|}
\hline
Data set &  No. of Instances & No. of Features &  No. of Classes\\
\hline
dermatology & 366 & 34 & 6\\
german & 1\,000 & 24 & 2\\
semeion & 1\,592 & 265 & 2\\
car & 1\,728 & 6 & 4\\
abalone & 4\,177 & 8 & 28\\
adult & 32\,561 & 14 & 2\\
bank-full & 45\,211 & 16 & 2\\
connect-4 & 67\,556 & 42 & 3\\
diabetes & 101\,766 & 49 & 3\\
census-income & 199\,523 & 41 & 2\\
\hline
\end{tabular}
\label{tab1}
\end{table*}

\subsection{Experiment I - Evaluation of Different Neuron Functions}

In this experiment, we tested the nine multivariate functions - sum, min, max, multiply, mean, std, var, median, and norm - across ten datasets (Table~\ref{tab1}) using a two-layer KAN architecture [n$_{in}$, 10, 1]. %This setup allows for exploration of 81 unique combinations of functions between the layers. Each dataset underwent testing with all 81 function settings, which were then ranked based on test performance. 
The average rank for the top ten functions combinations across all datasets is presented in Table~\ref{tab2}. 

The analysis revealed that mean and std were the top performers in the first layer, each appearing in five of the top 10 settings, while the second layer exhibited a more varied selection of functions. Table~\ref{tab3} focuses on scenarios where the same function is utilized in both layers. Intriguingly, mean emerged as the best-performing function, followed by std, with sum ranking third. Not surprisingly, the non-differentiable functions such as min, median, and max exhibited inconsistent performance. We decided to focus on using the mean function for the rest of our analysis as it consistently delivered the best performance based on the results. It is worth noting that the mean function is consistent with the Kolmogorov-Arnold representation theorem. This can be observed by directly modifying Equation~(\ref{kat}) using $\Phi'_q(x) = \frac{\Phi_q(x)}{2n+1}$ and $\phi'_{q,p}(x) = \frac{\phi_{q,p}(x)}{n}$, resulting in:
\begin{equation}
f(x_1, \ldots, x_n) = \frac{1}{2n+1}\sum_{q=1}^{2n+1} \Phi^{\prime}_q \left( \frac{1}{n}\sum_{p=1}^n \phi^{\prime}_{q,p}(x_p) \right)
\label{katavg}
\end{equation}
This simplifies to applying averaging to the scaled functions $\Phi^{\prime}_q(x)$, and $\phi^{\prime}_{q,p}(x)$ instead of using summation in the original formula involving $\Phi_q(x)$, and $\phi_{q,p}(x)$.

\begin{figure*}[t]
\centering
\subfloat{\includegraphics[width=1.\textwidth]{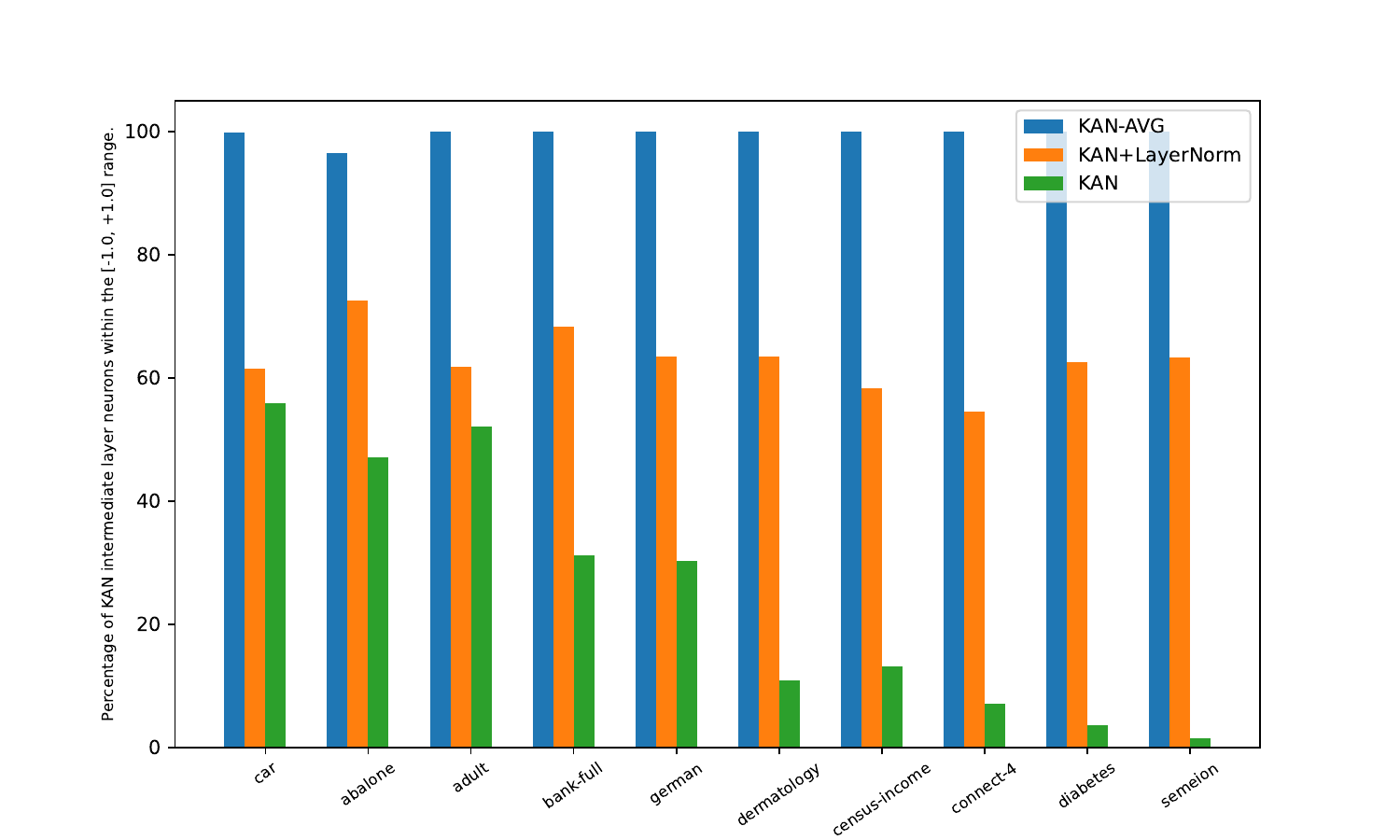}}
\caption{The percentage of KAN intermediate layer neurons' outputs within the effective grid range of [-1.0, +1.0], the datasets are ordered from left to right by ascending number of features.}
\label{fig1}
\end{figure*}

Our hypothesis posits that mean outperforms sum due to its ability to maintain input values within the effective range of the spline activation function. By default, the splines in KANs are defined with a range of [-1.0, +1.0], and values outside this range can lead to unpredictable activation behavior. The mean function mitigates this issue by reducing the output magnitude of neurons, increasing the likelihood that they fall within the desired range. This claim was validated through training KANs with both mean and sum functions in neurons, tracking the frequency with which neuron values fell within the specified range during training. As depicted in Figure ~\ref{fig1}, as the number of features increased, the standard KAN faces difficulty in keeping the values within the desired range in the intermediate layers. We found that the suggestion of \cite{li2024kolmogorov} to use Layer Normalization before the splines was effective in addressing covariate shift by centering neuron outputs around zero, but did not consistently ensure values fell within [-1.0, +1.0]. The adoption of mean in neurons resulted in exemplary performance, achieving consistent adherence to the desired range across datasets with 20 or more features. Even datasets with fewer features recorded values within the range over 99.0\% of the time, barring the '{\tt abalone}' dataset with a 96.51\% adherence rate.

%\subsubsection{Why is Mean doing better than sum?}

\begin{table*}[t]
\caption{The average rank of different settings of neuron functions, lower values of rank indicate higher performance across the 10 datasets.}
\begin{center}
\begin{tabular}{|c|c|c|}
\hline

\textbf{neuron function at layer 1} & \textbf{neuron function at layer 2} & \textbf{average rank} \\
\hline
mean & var & 14.9 $\pm15.6$ \\
\hline
mean & norm & 15.1 $\pm12.35$\\
\hline
std & sum & 15.3 $\pm16.61$\\
\hline
mean & sum & 17.2 $\pm16.21$\\
\hline
mean & std & 17.6 $\pm14.25$\\
\hline
std & var & 18.3 $\pm11.63$\\
\hline
mean & mean & 19.1 $\pm14.11$\\
\hline
std & mean & 19.4 $\pm17.26$\\
\hline
std & std & 20.6 $\pm17.26$\\
\hline
std & norm & 20.6 $\pm12.92$\\
\hline
\end{tabular}
\label{tab2}
\end{center}
\end{table*}

\begin{table*}[t]
\caption{The average rank of different settings when the same neuron function is used in both KAN layers, lower values of rank indicate higher performance across the 10 datasets.}
\begin{center}
\begin{tabular}{|c|c|c|}
\hline

\textbf{neuron function at layer 1} & \textbf{neuron function at layer 2} & \textbf{average rank} \\
\hline
mean & mean & 19.1 $\pm14.11$ \\
\hline
std & std & 20.6 $\pm17.26$\\
\hline
sum & sum & 25.7 $\pm18.26$\\
\hline
norm & norm & 26.3 $\pm17.15$\\
\hline
var & var & 28.5 $\pm21.69$\\
\hline
min & min & 48.5 $\pm24.45$\\
\hline
median & median & 57.1 $\pm18.99$\\
\hline
max & max & 60.5 $\pm23.73$\\
\hline
multiply & multiply & 64.1 $\pm15.42$\\
\hline
\end{tabular}
\label{tab3}
\end{center}
\end{table*}

\subsection{Experiment II - Mean vs Sum as a Neuron Function}
In this experiment, we will compare the performance of three different variants of KAN. The first variant, denoted as KAN, represents the standard KAN architecture with summation operation at the neurons. The second variant, denoted as KAN+LayerNorm, incorporates layer normalization in the intermediate layers of the standard KAN architecture. The third variant, referred to as KAN-AVG, utilizes mean function within the neurons of the KAN architecture. We will conduct 20 independent training runs for each dataset using each variant of KAN. To compare the average test accuracy of these variants, we will employ the Wilcoxon signed-rank test for statistical significance assessment.

The advantage of utilizing the mean over the sum in KAN neurons is highlighted in Table~\ref{tab4}, where KAN-AVG consistently demonstrates higher average test accuracy across all datasets compared to the standard KAN. Furthermore, KAN-AVG outperforms the standard KAN significantly in terms of test accuracy in 7 datasets, as confirmed by Wilcoxon test results. It is worth noting that the performance variability of KAN-AVG is consistently lower than that of the standard KAN, indicating a more stable training process. 

Additionally, the incorporation of layer normalization positively impacts KAN, resulting in significantly higher average test accuracy on four occasions. However, despite these improvements, KAN-AVG still surpasses KAN+LayerNorm in terms of average test accuracy, with KAN-AVG recording signficnatly higher average test accuracy four times compared to KAN+LayerNorm.

\begin{table*}[t]
\caption{Results of KAN networks trained with KAN, KAN+LayerNorm, and KAN-AVG. Each result is the mean of 20 independent runs; bold indicates best results, single asterisk (*) indicates statistical significance against KAN, and double asterisk (**) indicates statistical significance against both KAN and KAN+LayerNorm.}
\begin{center}
\begin{tabular}{|c|c|c|c|}
\hline
\textbf{Dataset} & \textbf{KAN} & \textbf{KAN+LayerNorm} & \textbf{KAN-AVG} \\
\hline
dermatology & 75.34\%\,$\pm$\,5.67 & 90.74\%\,$\pm$\,2.61* & \textbf{94.05\%\,$\pm$\,1.31**} \\
\hline
german & 66.3\%\,$\pm$\,3.15 & 65.12\%\,$\pm$\,3.54 & \textbf{71.05\%\,$\pm$\,2.11**} \\
\hline
semeion & 93.62\%\,$\pm$\,4.86 & \textbf{98.29\%\,$\pm$\,0.54} & 97.71\%\,$\pm$\,0.39 \\
\hline
car & 90.38\%\,$\pm$\,2.08 & \textbf{92.96\%\,$\pm$\,1.37*} & 92.33\%\,$\pm$\,1.46* \\
\hline
abalone & 25.44\%\,$\pm$\,1.23 & 25.74\%\,$\pm$\,1.31 & \textbf{26.78\%\,$\pm$\,1.23**} \\
\hline
adult & 84.84\%\,$\pm$\,0.25 & 84.8\%\,$\pm$\,0.24 & \textbf{85.03\%\,$\pm$\,0.15**} \\
\hline
bank-full & 90.21\%\,$\pm$\,0.32 & 90.29\%\,$\pm$\,0.28 & \textbf{90.38\%\,$\pm$\,0.17} \\
\hline
connect-4 & 68.19\%\,$\pm$\,2.75 & \textbf{70.75\%\,$\pm$\,3.06*} & 70.27\%\,$\pm$\,1.58* \\
\hline
diabetes & 53.09\%\,$\pm$\,2.18 & 53.68\%\,$\pm$\,2.5 & \textbf{54.05\%\,$\pm$\,2.18} \\
\hline
census-income & 94.79\%\,$\pm$\,0.25 & 94.98\%\,$\pm$\,0.12* & \textbf{95.04\%\,$\pm$\,0.12*} \\
\hline
\end{tabular}
\label{tab4}
\end{center}
\end{table*}

%In this experiment, we compare KANs equipped with DropKAN layers against KANs with standard KAN layers and KANs regularized using Dropout. The goal of the experiment is to validate the regularizing effect of DropKAN and to compare it against using standard Dropout with KANs. 

%We use a random search to optimize the rates of drop for the DropKAN and Dropout settings, we ran the search for 50 evaluation per setting, and choose the setting with the highest accuracy performance on the validation split. For evaluation we train every setting five times and report the average of the runs. 

%The results in Table~\ref{tab4} suggest that regularization methods improved the standard KAN performance (No-Drop) in 8 scenarios apart from two datasets {\tt semeion}, and {\tt car}. DropKAN w/ scale is clearly the most successful setting recording the highest test accuracy in six of the ten datasets, while the standard Dropout w/ scale was only the best choice in one dataset.   

%The results suggest that scaling is showing a positive effect with DopKAN as the scaled settings performed better than w/o scaling in seven scenarios, consistent with the results of the previous experiment and Equation~\ref{dropkaneq}. However, scaling is not as successful in the case of Dropout showing better performance comparing to no scaling in only 4 scenarios.   

\section{Related Work}

The original KAN paper by Liu et al. \cite{liu2024kan} demonstrated the superior performance of KANs over MLPs in data fitting and PDE tasks while utilizing fewer parameters. Subsequent studies have further validated the efficacy of KANs across various domains such as computer vision \cite{cheon2024demonstrating,azam2024suitability}, time series analysis \cite{vaca2024kolmogorov}, tabular data \cite{poeta2024benchmarking}, engineering design \cite{abueidda2024deepokan}, human activity recognition \cite{liu2024ikan,liu2024initial}, DNA sequence prediction \cite{he2024kan}, and quantum architecture search \cite{kundu2024kanqas}. However, concerns regarding the robustness of KANs to noise have been raised, as evidenced by the degradation in performance when noise is introduced \cite{shen2024reduced}. Additionally, Le et al. \cite{le2024exploring} reported that KANs failed to overcome MLPs in terms of performance while requiring substantially higher hardware resources. Another study \cite{yu2024kan} reported contrasting results, comparing KANs and MLPs models across various tasks, suggesting that MLP models generally outperformed KANs, except in tasks related to symbolic formula representation, where the advantage of KANs stemmed mainly from their distinctive B-spline activation function.

Moreover, research has expanded beyond the standard KAN architecture to explore new variations and enhancements, such as graph-based architectures \cite{bresson2024kagnns,kiamari2024gkan,de2024kolmogorov,xu2024fourierkan}, convolutional KANs \cite{bodner2024convolutional}, and transformer-based KANs \cite{genet2024temporal}. Furthermore, ideas originating from deep learning, such as DropKAN \cite{altarabichi2024dropkan} improved the generalization of KANs by embedding a dropout mask within the KAN layer. Further efforts to enhance the design and efficiency of KANs focused on exploring alternatives to B-spline to represent activation functions, such as wavelets \cite{bozorgasl2024wav,seydi2024unveiling}, radial basis functions \cite{li2024kolmogorov,ta2024bsrbf}, fractional functions \cite{aghaei2024fkan}, rational functions \cite{aghaei2024rkan}, and sinusoidal functions \cite{reinhardt2024sinekan}. Notably, the concept of utilizing the mean instead of the sum in KAN neurons, as proposed in this study, could be easily extended to such architectures and custom implementations with minimal implementation overhead.

\section{Conclusions}
In this paper, we suggest replacing the summation in KAN neurons with an averaging function. Our experiments show that employing the average function results in more stable training, ensuring that the inputs remain within the effective range of the spline activations. Utilizing the average function clearly aligns with the Kolmogorov-Arnold representation theorem, as illustrated straightforwardly.

\bibliography{references}  %%% Remove comment to use the external .bib file (using bibtex).

\end{document}